\documentclass[onefignum,onetabnum]{siamonline171218}

\usepackage{amssymb,amsmath,amsfonts,mathtools}

\usepackage{graphicx,rotate}
\graphicspath{{./Figures/}}
\usepackage{wrapfig}

\usepackage[mathscr]{euscript}
\usepackage{caption}

\usepackage{tikz}
\usepackage{subcaption}
\usepackage[export]{adjustbox}
\usepackage{dsfont}
\usepackage{bigstrut}
\usepackage{multirow}
\usepackage{setspace}
\usepackage{tikz}
\usepackage[algo2e,linesnumbered,lined,noend]{algorithm2e}
\usepackage{ulem}
\usepackage{mathabx}
\usepackage{bm}
\usepackage{booktabs}

\usepackage{enumitem}
\setlist[enumerate]{leftmargin=.5in}
\setlist[itemize]{leftmargin=.5in}

\newsiamthm{assumption}{Assumption}

\newsiamthm{remark}{Remark}

\def\X{\mathcal{X}}
\def\Ri{\mathcal{R}_i}

\def\bu{{\bm{u}}}

\def\U{\mathcal{U}}
\def\Ui{\mathcal{U}_i}
\def\Uii{\mathcal{U}_{-i}}

\def\bpi{\bm{\pi}}

\def\bmu{\bm{\mu}}
\def\pist{\pi^{\ast}}
\def\bpist{\bpi^{\ast}}

\def\bR{\bm{\mathcal{R}}}
\def\bU{\bm{\mathcal{U}}}
\def\br{\bm{r}}

\def\bV{\bm{V}}
\def\Vhat{\hat{V}}

\def\bQ{\bm{Q}}

\def\Qhat{\hat{Q}}

\def\D{\mathcal{D}}
\def\Y{\mathcal{Y}}

\DeclareBoldMathCommand\bQhat{\Qhat}
\DeclareBoldMathCommand\bVhat{\Vhat}

\def\br{\bm{r}}
\def\bpsi{\bm{\psi}}
\def\bPsi{\bm{\Psi}}
\def\bP{\bm{P}}
\def\NN{{\mathds{N}}}

\def\N{\bm{\mathcal{N}}}
\def\mvn{{N}}

\newcommand*{\txtNashOp}[1]{\N_{\vspace{1em}\hspace{-0.3em}\scriptscriptstyle{} #1}}
\newcommand*{\NashOp}[1]{\mathchoice{\underset{#1}{\N}}{\txtNashOp{#1}}{\txtNashOp{#1}}{\txtNashOp{#1}}}

\newcommand{\EE}[1]{\underset{{\scriptscriptstyle #1}}{\E}}

\def\Ahat{\widehat{A}}

\def\bAhat{\bm{\Ahat}}

\def\Lhat{\hat{\mathcal{L}}}
\def\L{\mathcal{L}}

\def\bnu{\bm{\nu}}
\def\nubar{\bar{\nu}}

\def\Y{\mathcal{Y}}
\def\D{\mathcal{D}}
\def\R{\mathds{R}}
\def\E{\mathbb{E}}

\def\mfN{{\mathfrak{N}}}
\def\mfT{\mathfrak{T}}

\usepackage{marginnote}
\definecolor{WowColor}{rgb}{.75,0,.75}
\definecolor{SubtleColor}{rgb}{0,0,.50}

\newcounter{margincounter}

\SetKwInOut{Input}{Input}
\SetKwInOut{Output}{Output\,}
\SetKwInOut{Initialize}{Initialize}
\SetKwInOut{Reset}{Reset}
\SetKwInOut{Data}{Data}

\title{Deep Q-Learning for Nash Equilibria: Nash-DQN\thanks{SJ would like to acknowledge the support of the Natural Sciences and Engineering Research Council of Canada
(NSERC), [funding reference numbers RGPIN-2018-05705 and RGPAS-2018-522715].}}

\author{
Philippe Casgrain\thanks{Deparment Mathematics, ETH Zurich (\email{Philippe Casgrain <philippe.casgrain@math.ethz.ch>}).}
\and
Brian Ning \thanks{Department of Statistical Sciences, University of Toronto, Canada (\email{brian.ning@mail.utoronto.ca})}.
\and
Sebastian Jaimungal\thanks{Department of Statistical Sciences, University of Toronto, Canada (\email{sebastian.jaimungal@utoronto.ca}, \url{http://sebastian.statistics.utoronto.ca.})}
}

\begin{document}

\maketitle

\begin{abstract}
Model-free learning for multi-agent stochastic games is an active area of research. Existing reinforcement learning algorithms, however,  are often restricted to zero-sum games, and are applicable only in small state-action spaces or other simplified settings. 
Here, we develop a new data efficient Deep-Q-learning methodology for model-free learning of Nash equilibria for general-sum stochastic games. The algorithm uses a local linear-quadratic expansion of the stochastic game, which leads to analytically solvable optimal actions. The expansion is parametrized by deep neural networks to give it sufficient flexibility to learn the environment without the need to experience all state-action pairs. We study symmetry properties of the algorithm stemming from label-invariant stochastic games and as a proof of concept, apply our algorithm to learning optimal trading strategies in competitive electronic markets.
\end{abstract}

\section{Introduction}

The study of equilibria in systems of interacting agents is ubiquitous throughout the natural and social sciences. The classical approach to studying these equilibria requires building a model of the interacting system, solving for its equilibrium, and studying its properties thereafter. This approach often runs into complications, however, as a fine balance  between (i) model tractability and (ii) its ability to capture the main features of the data it aims to represent, must be struck. Rather than taking a model-based approach, it is possible to derive non-parametric reinforcement-learning (RL) methods to study these equilibria. The main idea behind  these methods is to directly approximate equilibria from simulations or observed data, providing a powerful alternative to the usual approach.

The majority of the existing literature on RL is dedicated to single-player games. Most modern approaches follow either a deep Q-learning approach(e.g.~\cite{mnih2013playing}), policy gradient methods (e.g.~\cite{sutton2000policy}), or some mixture thereof (e.g.~\cite{hessel2018rainbow}). RL  methods have also been developed for multi-agent games, but are for the most part restricted to the case of zero-sum games. For a survey see~\cite{bu2008comprehensive}. 

There are recent efforts on extending RL to general sum games with fictitious play as in~\cite{hu2019deep}, or with iterative fixed point methods as in~\cite{lanctot2017unified}. In the specific context of (discrete state-action space) mean-field games, \cite{guo2019learning} provides a Q-learning algorithm for solving for the Nash-equilibria. Many of the existing algorithms suffer either from computational intractability as the size and complexity of a game increases, when state-action space becomes continuous, or from the inability to model complex game behaviour.

Hu and Wellman~\cite{hu2003nash} introduce a Q-learning based approach for obtaining Nash equilibria in general-sum stochastic games. Although they prove convergence of the algorithm for games with finite game and action spaces,  their approach is computationally infeasible for all but the simplest examples. The main computational bottleneck in their approach is the need to repeatedly compute a local Nash equilibrium over states, which is an NP-hard operation in general. Moreover, the method proposed in \cite{hu2003nash} does not extend to games where agents choose continuous-valued controls or to games with either high-dimensional game state representations or with large numbers of players. We instead combine the \textit{iLQG} framework of of~\cite{todorov2005generalized,gu2016continuous} and the Nash Q-learning algorithm of~\cite{hu2003nash} to produce an algorithm which can learn Nash equilibria in these more complex and practically relevant settings. 

In particular, we decompose the state-action value (Q)-function as a sum of the value function and the \textit{advantage} function. We approximate the value function using a neural-net, and we locally approximate the advantage function as linear-quadratic in the agents' actions with coefficients that are non-linear functions of the features given by a neural-net. This allows us to compute the Nash equilibrium analytically at each point in feature space (i.e., the optimal action of all agents) in terms of the network parameters. Using this closed form local Nash equilibrium, we derive an iterative actor-critic algorithm to learn the network parameters.

In principle, our approach allows us to deal with stochastic games with a large number of game state features and a large action space. Moreover, our approach can be easily adapted to mean-field game (MFG) problems, which result from the infinite population limit of certain stochastic games (see \cite{huang2006large,lasry2007mean,carmona2017probabilistic}), such as those developed in, e.g., \cite{casgrain2020mean,casgrain2018algorithmic} or major-minor agent MFGs such as those studied in, e.g., \cite{huang2010large,nourian2013ep,huang2015mean}. A drawback of the method we propose is the restriction on the local structure of the proposed Q-function approximator. We find, however, that the proposed approximators are sufficiently expressive in most cases, and perform well in the numerical examples that we include in this paper.

The remainder of this paper is structured as follows. \Cref{sec:Model-Setup} introduces a generic Markov model for a general-sum stochastic game. In \Cref{sec:Optimality-Conditions}, we present optimality conditions for the stochastic game and motivate our Q-learning approach to finding Nash equilibria. \Cref{sec:Locally-Linear-Quadratic-Algortithm} introduces our local linear-quadratic approximations to the Q-function and the resulting learning algorithm. We also provide several simplifications that arise in label-invariant games. \Cref{sec:Implementation-Details-Section} covers implementation details and \Cref{sec:Experiments-Section} presents some illustrative examples. 

\section{Model Setup}
\label{sec:Model-Setup}

We consider a stochastic game with  agents $i\in\mfN:=\{1 , 2 , \dots , N\}$ all competing together. We assume  the state of the game is represented via the stochastic process $\{x_t\}_{t\in\NN}$ so that for each time $t$, $x_t \in \X$, for some separable Banach space $\X$. At each time $t$, agent-$i$ chooses an action $u_{i,t}\in\Ui$, where $\Ui$ is assumed to be a separable Banach space. In the sequel, we use the notation $\bu_{-i,t}=(u_{j,t})_{ j\in\mfN/\{i\}}$ to denote the vector of actions of all agents other than agent-$i$ at time $t$ and the notation $\bu_{t}=(u_{j,t})_{ j\in\mfN}$ to denote the vector of actions of all agents. We assume that the game is a Markov Decision Process (MDP) with a fully-visible game state. The MDP assumption is equivalent to assuming  the joint state-action process $\left(x_t, \bu_t\right)_{t=1}^{\infty}$ is Markov, whose state transition probabilities are defined by the stationary Markov transition kernel $p(x_{t+1} \mid x_t, \bu_t)$ and a distribution over initial states $p_0(x_0)$.

At each step of the game, agents receive a reward that varies according to the current state of the game, their own choice of actions, and the actions of all other agents. The agent-$i$'s reward is represented by the function $(x,u_i,\bu_{-i})\mapsto r_i(x;u_i,\bu_{-i})\in\R$, so that at each time $t$, agent-$i$ accumulates a reward $r_i(x_t;u_{i,t},\bu_{-i,t})$. We assume that each function $r_i$ is continuously differentiable and concave in $u_i$ and is continuous in $x$ and $\bu_{-i}$.

At each time $t$, agent-$i$ may observe other agents' actions $\bu_{-i,t}$, as well as the state of the game $x_t$. Moreover, each agent-$i$ chooses their actions according to a deterministic Markov policy $ \X \ni x \mapsto \pi_i(x) \in \Ui$. The objective of agent-$i$ is to select the  policy $\pi_i$ that maximizes the objective functional $\Ri$ which represents their personal expected discounted future reward over the remaining course of the game, given a fixed policy $\pi_i$ for themselves and a fixed policy $\bpi_{-i}$ for all other players. The objective functional for agent-$i$ is
\begin{equation} \label{eq:Agent-Objective}
	\Ri(x;\pi_i,\bpi_{-i}) =
	\E\left[ \sum_{t=0}^\infty \gamma_i^{-t} \, r_{i}(x_t,\pi_{i,t},\bpi_{-i,t}) \right],
\end{equation}
where the expectation is over the process $(x_t)_{t\in\NN}$, with $x_0=x$, and where we assume $\gamma_i \in (0,1)$ is a fixed constant representing a discount rate. In Equation~\eqref{eq:Agent-Objective}, we use the compressed notation $\pi_{i,t} := \pi_i(x_t)$ and $\bpi_{-i,t} := \bpi_{-i}(x_t)$.  The agent's objective functional~\eqref{eq:Agent-Objective} explicitly depends on the policy choice of all agents. Each agent, however, can only control their own policy, and must choose their actions while conditioning on the behavior of all other players.

Agent-$i$ therefore seeks a policy that optimizes their objective function, but remains robust to the actions of others. In the end, agents' policies form a Nash equilibrium -- a collection of policies $\bpist(x) = \{ \bpist_i(x) \}_{i\in\mfN}$ such that unilateral deviation from this equilibrium by a single agent will result in a decrease in the value of that agent's  objective functional. Formally, we say that a collection of policies $\bpist$ forms a Nash equilibrium if \begin{equation} \label{eq:Nash-equilibrium-def}
	\Ri\left( x ; \pi_i , \bpist_{-i} \right)
	\leq
	\Ri\left( x ; \pist_i , \bpist_{-i} \right)
\end{equation}
for all admissible policies $\pi_i$ and for all $i\in\mfN$.
Informally, we can interpret the Nash equilibrium as the policies for which each agent simultaneously maximizes their own objective function, while conditioned on the actions of others.

\section{Optimality Conditions}
\label{sec:Optimality-Conditions}

Our ultimate goal is to obtain an algorithm that can attain the Nash equilibrium of the game without a-priori knowledge of its dynamics. In order to do this, we first identify conditions that are more easily verifiable than the formal definition of a Nash  equilibrium given above.

We proceed by extending the well known Bellman equation for Nash equilibria. While leaving $\bpist_{-i}$ fixed, we may apply the dynamic programming principle to agents-$i$ reward $\Ri(x;\pist_i,\bpist_{-i})$ resulting in
\begin{equation} \label{eq:Bellman-Eq}
	\Ri(x;\pist_i,\bpist_{-i})
	=
	\max_{u\in\Ui} \left\{
	r_i(x,u,\bpist_{-i}(x)) + \gamma_i \,  \EE{x^\prime \sim p(\cdot \mid x,\bu)} \left[ \Ri(x^\prime;\pist_{i,1},\bpist_{-i,1}) \right]
	\right\}.
\end{equation}
At the Nash equilibrium, equation~\eqref{eq:Bellman-Eq} is satisfied simultaneously for all $i\in\mfN$.

To express this more concisely, we introduce a vector notation. First define the vector-valued function $\bR(x;\bpi)=(\Ri(x;\pi_i,\bpi_{-i}))_{i\in\mfN}$, consisting of the stacked vector of objective functions. We call the stacked objective functions evaluated at their Nash equilibria the stacked value function, which we write as $\bV(x):=\left(V_i(x)\right)_{i\mfN} = \bR(x;\bpist)$.
 % Using this notation, we introduce the \textit{agent-$i$Q$-function} as follows.

Next, we define the Nash state-action value function, also called the Q-Function, which we denote $\,\bQ(x;\bu) := \left( Q_i(x;u_i,\bu_{-i})\right)_{i\in\mfN}$, where
\begin{equation}
	\bQ(x;\bu) = \br(x;\bu) + \gamma_i  \EE{x^\prime \sim p(\cdot \mid x,\bu)} \left[ \bV(x^\prime) \right]
	\;,
\end{equation}
and where we denote $\br(x,\bu):=\left(r_i(x,u_{i},\bu_{-i})\right)_{i\in\mfN}$ to indicate the vectorized reward function.
Each element $Q_i$ of $\bQ$ can be interpreted as the expected maximum value their objective function may take, given a fixed current state $x$ and a fixed (arbitrary) immediate action $\bu$ taken by all agents.

% \TBD{Each element $Q_i$ of $\bQ$ can be interpreted as an agent taking an arbitrary action $\bu$ in state $x$ at the "current point in time", and then all agents subsequently following the optimal action.}

Next, we define the Nash operator as follows.
\begin{definition}[Nash Operator]
	Consider a collection of $N$ concave real-valued functions, $\bm{f}(\bu) = (f_i(u_i, \bu_{-i}))_{i\in\mfN}$, where $f_i: \Ui \bigtimes \Uii \rightarrow \R$. We define the Nash operator $\NashOp{\bu\in\bU} \bm{f}(\bu) \mapsto \bm{f}(\bu^{\ast})$, as a map from the collection of functions $\bf$ to their Nash equilibrium value $\bu^{\ast}\in\R^{N}$, where $\bu^{\ast} = \arg \NashOp{\bu\in\bU} \bm{f}(\bu)$ is the unique point satisfying,
	\begin{align}
		f_i\left( \omega_i, \bu_{-i}^\ast  \right)
		\leq f_i\left( u_i, \bu_{-i}^\ast  \right),
		\qquad \forall \omega_i \in \Ui, \text{ and } \forall i\in\mfN.
	\end{align}
\end{definition}
For a sufficiently regular collection of functions $\bm{f}$, the Nash operator corresponds to simultaneously maximizing each of the $f_i$ in their first argument $u_i$.

This definition provides us with a relationship between the value function and  agents' $Q$-function as  $\bV(x) = \NashOp{\bu\in\bU} \bQ(x;\bu)$. Using the Nash operator, we may then express the Bellman Equation~\eqref{eq:Bellman-Eq} in a concise form as
\begin{align} \label{eq:Nash-Op-Relationship}
	\bV(x) &= \NashOp{\bu\in\bU} \bQ(x;\bu)
	%\\ &
	=
	\NashOp{\bu\in \bU}
	\left\{
		\br(x;\bu) + \gamma_i \, \EE{x^\prime \sim p(\cdot \mid x,\bu)}\left[ \bV(x^{\prime}) \right]
	\right\}
	% \\ &=
	% \NashOp{\bu\in \bU}
	% \left\{
	% 	\br(x;\bu) + \gamma_i \, \EE{x^\prime \sim p(\cdot \mid x,\bu)}\left[ \NashOp{\bu^\prime \in \bU} \bQ(x^\prime;\bu^\prime) \right]
	% \right\}
	,
\end{align}
% \PROBLEM{PC: When is this condition uniquely satisfied? Maybe this is obvious or something? I probably need a reference, but I am not sure where to look and so on. More precisely: Is (3.5-3.6) a necessary and sufficient condition for a agent-$i$Q$-function?
% It feels like it should be the case. In addition, demonstrating this in a single player game setting should be enough.}
which we refer to as the \textit{Nash-Bellman equation} for the remainder of the paper.
The definition of the value function equation~\eqref{eq:Nash-Op-Relationship} implies that $\bpist = \NashOp{\bu\in\bU} \bQ(x;\bu)$. Hence, in order to identify the Nash equilibrium $\bpist$, it is sufficient to obtain the $Q$-function and apply the Nash operator to it. This principle will inform the approach we take in the remainder of the paper: rather than directly searching the space of policy collections for the Nash-equilibrium via equations~\eqref{eq:Agent-Objective} and~\eqref{eq:Nash-equilibrium-def}, we may rely on identifying the function $\bQ$ satisfying \eqref{eq:Nash-Op-Relationship}, and thereafter compute $\bpist = \NashOp{\bu\in\bU} \bQ(x;\bu)$.

\section{Locally Linear-Quadratic Nash Q-Learning}
\label{sec:Locally-Linear-Quadratic-Algortithm}

In this section, we formulate an algorithm which learns the Nash equilibrium of the stochastic game described in the previous section. The principal idea behind the approach we take is to construct a parametric estimator $\bQhat^\theta$ of  agent-$i$'s $Q$-function, where we search for the set of parameters $\theta\in\Theta$, which results in estimators $\bQhat^\theta$ that approximately satisfy the Nash-Bellman equation~\eqref{eq:Nash-Op-Relationship}. Thus, our objective is to minimize the quantity
\begin{equation} \label{eq:Expected-Value-Nash-Gap}
	\EE{\substack{ \vphantom{A}  x \sim \rho \\ x^\prime \sim p(\cdot \mid x,\bu)}}
	\left[ \; \left\|
		\bQhat^\theta ( x ; \bu )
		- \br(x;\bu) - \gamma_i \, \NashOp{\bu^\prime \in\U} \bQhat^\theta(x^\prime; \bu^\prime )
	\right\|^2 \; \right]
	\;,
\end{equation}
over all $u$, where we define $\rho$ to be an unconditional proability measure over game states $x$. Equation~\eqref{eq:Expected-Value-Nash-Gap} is designed as a measure of the gap between the right and left sides of equation~\eqref{eq:Nash-Op-Relationship}. We may also interpret it as the distance between $\bQhat^\theta$ and the true value of $\bQ$. The expression~\eqref{eq:Expected-Value-Nash-Gap} is intractable, since we do not know $\rho$ nor $p(\cdot|x,\bu)$ a-priori, and we wish to make little to no assumptions on the system dynamics. Therefore, we rely on a simulation based method  and approximate  \eqref{eq:Expected-Value-Nash-Gap} with
\begin{equation} \label{eq:Empirical-Nash-Gap}
	\L(\theta) =\frac{1}{M}\sum_{m=1}^M
	\left\lVert
		\bQhat^\theta ( x_{m} ; \bu_{m} )
		- \br(x_{m};\bu_{m}) - \gamma_i \, \NashOp{\bu_{m}^\prime \in\U} \bQhat^\theta(x_{m}^\prime; \bu_{m}^\prime )
	\right\rVert^2
	\;,
\end{equation}
where for each $m=1,2,\dots,M$, $(x_m,\bu_m;x^\prime_m)$ represents an observed transition triplet from the game. We then   search for $\theta\in\Theta$ that minimizes the $\L(\theta)$ in order to approximate $\bQhat$.

Our approach is motivated by Hu and Wellman~\cite{hu2003nash} and Todorov \& Li~\cite{todorov2005generalized}. \cite{hu2003nash} presents a $Q$-learning algorithm where $\bQhat$, which is assumed to take only finitely many values, can be estimated through an update rule that relies on the repeated computation of the Nash operator $\NashOp{\bu\in\U} \bQhat^\theta$. As the computation of $\NashOp{\bu\in\U} \bQhat^\theta$ is NP-hard in general, this approach proves to be computationally intractable beyond trivial examples. To circumvent this issue and to make use of more expressive parametric models, we generalize and adapt techniques in Gu et al.~\cite{gu2016continuous} to the multi-agent game setting to develop a computational and data efficient algorithm for approximating the Nash equilibria.

In our algorithm, we make the additional assumption that game states $x\in\X$ and actions $u_i\in\U_i$ are real-valued. Specifically, we assume that $\X=\R^{d_x}$ for some positive integer $d_x$ and $\Ui=\R^{d_i}$ for each $i\in\mfN$, where $d_1,\dots d_N$ are all positive integers. For notational convenience we define $d_{-i}:=\sum_{j\in\mfN/\{i\}} d_j$.\footnote{Our approach  can be easily extended to the case of controls that are restricted to convex subsets of $\R^{d_i}$.}

We now define a specific model for the collection of approximate $Q$-functions $\bQhat^{\theta}(x;\bu) = ( \Qhat^\theta_i(x;\bu) )_{i\in\mfN}$. For each $\theta\in\Theta$, we have $\bQhat^{\theta}:\X\bigtimes\bU \rightarrow \R^N$ and decompose the $Q$-function into two components:
\begin{equation} \label{eq:Q-Function-Decomposition}
	\bQhat^{\theta}(x;\bu) = \bVhat^\theta(x) + \bAhat^\theta (x;\bu)\,,
\end{equation}
where $\bVhat^\theta(x) = ( \Vhat^\theta_i(x) )_{i\in\mfN}$ is a model of the collection of value functions so that $\bVhat^{\theta} : \X \rightarrow \R^N$ and where  $\bAhat^\theta (x;\bu) = ( \Ahat_i^\theta (x;\bu) )_{i\in\mfN}$ is what we refer to as the collection of \textit{advantage functions}. The advantage function represents the optimality gap between $\bQhat$ and $\bVhat$. We  further assume that for each $i\in\mfN$, $\Ahat^\theta_i$ has the linear quadratic form
\begin{equation}\label{eq:LQ-Advantage}
\begin{aligned}
	{\Ahat}_{i}^{\theta} (x;\bu) =
	&-
	\begin{pmatrix}
		u_i - \mu^\theta_i(x) \\
		\bu_{-i} - \bmu^\theta_{-i}(x)
	\end{pmatrix}^{\intercal} \bP_i^\theta(x)
	\begin{pmatrix}
		u_i - \mu^\theta_i(x) \\
		\bu_{-i} - \bmu^\theta_{-i}(x)
	\end{pmatrix}
	%\\ &
	+
	\begin{pmatrix}
		\bu_{-i} - \bmu^\theta_{-i}(x)
	\end{pmatrix}^\intercal
	\bPsi_i^{\theta}(x)
	\;,
\end{aligned}
\end{equation}
where the block matrix
\begin{equation}
	\bP_i^{\theta}(x) := \begin{pmatrix}
		P_{11,i}^\theta(x) & P_{12,i}^\theta(x) \\ P_{21,i}^\theta(x) & P_{22,i}^\theta(x)
	\end{pmatrix},
\label{eqn:PblockDef}
\end{equation}
$\mu^{\theta}_i(x) : \X \rightarrow \R^{d_i}$ with $\bmu^{\theta}_{-i}(x) = (\mu^{\theta}_i(x))_{i\in\mfN}$, and  $\bPsi_i:\X\rightarrow\R^{d_{-i}}$.
In \eqref{eqn:PblockDef},  $P^{\theta}_{11,i} : \X \rightarrow \R^{d_i \times d_i}$,  $P^{\theta}_{12,i} : \X \rightarrow \R^{d_{i} \times d_{-i}}$, $P^{\theta}_{21,i} : \X \rightarrow \R^{d_{-i} \times d_{i}}$ and $P^{\theta}_{22,i} : \X \rightarrow \R^{d_{-i} \times d_{-i}}$ are matrix valued functions, for each $i\in\mfN$. We require that  $P^{\theta}_{11,i}(x)$ is positive-definite for all $x\in\X$ and without loss of generality we may choose $P_{12,i}(x) = \left(P_{21,i}(x)\right)^\intercal$, as the advantage function depends only the symmetric combination of $P_{12,i}$ and $P_{21,i}$.

Hence, rather than modelling $\bQhat^\theta (x;\bu)$, we instead model the functions $\bVhat^\theta$, $\bmu^{\theta}$, and $\{\bP^\theta_{i}, \bPsi_i^{\theta} \}_{i\in\mfN}$ separately as functions of the state space $\X$. Each of these functions can be modeled by  universal function approximators such as neural networks. The only major restriction is that  $P_{11,i}^{\theta}(x)$ must remain a positive-definite function of $x$. This restriction is easily attained by decomposing $P_{11,i}^{\theta}(x)$ using  Cholesky decomposition, so that we write $P_{11,i}^{\theta}(x) = L_{11,i}^{\theta}(x) (L_{11,i}^{\theta}(x))^{\intercal}$ and instead model the lower triangular matrices $L_{11,i}^{\theta}\in\R^{d_{i}\times d_{i}}$.

The model assumption in~\eqref{eq:LQ-Advantage} implicitly assumes that agent-$i$'s $Q$-function  can be approximately written as a linear-quadratic function of the actions of each agent. One can equivalently motivate such an approximation by considering a second order Taylor expansion of  $Q_i$ in the variable $\bu$ around the Nash equilibrium, together with the assumption that the $Q_i$ are convex functions of their input $u_i$. This expansion, however, assumes nothing about the dependence of  $Q_i$ on the value of game state $x$.

The form of~\eqref{eq:LQ-Advantage} is designed so that each $\Qhat_i^{\theta}(x;u_i,\bu_{-i})$ is a concave function of $u_i$, guaranteeing that $\NashOp{u\in\bU}\bQhat$ is bijective. Moreover, under our model assumption, the Nash-equilibrium is attained at the point $u^*(x) = \bmu(x)$ and at this point, the advantage function is zero, hence we obtain simple expressions for the value function and the equilibrium strategy
\begin{equation}
	\bVhat^\theta(x) = \NashOp{u\in\bU} \bQhat^{\theta}(x;\bu)
	\quad \text{and} \quad
	\bmu(x) = \arg \NashOp{u\in\bU} \bQhat^{\theta}(x;\bu)\,.
\end{equation}
Consequently, our model allows us to directly specify the Nash equilibrium strategy and the value function of each agent through the functions $\bmu^\theta$ and $\bVhat^\theta$. The outcome of this simplification is that the summand of the loss function in equation~\eqref{eq:Empirical-Nash-Gap}, which contains the Nash equilibria and was itself previously intractable, becomes tractable. For each sample observation $m$ (consisting of a state $x_m$, an $\bu_m$, and new state $x'_m$) we then have a loss of
\begin{subequations}
\begin{equation}
	\L_m(\theta)= \left\lVert
		\bVhat^\theta(x_m) +
		\bAhat^\theta ( x_{m} ; \bu_{m} )
		- \br(x_{m};\bu_{m}) - \gamma_i \, \bVhat^\theta(x_m^{\prime})
	\right\rVert^2,
\end{equation}
and all that remains is to minimize the total loss
\begin{equation}
\L(\theta)=\tfrac{1}{M}\sum_{m=1}^M\L_m(\theta)
\end{equation}%
\label{eqn:LossLocallyLQNash}
\end{subequations}%
over the parameters $\theta$ given a set of observed state-action triples $(x_m,\bu_m,x'_m)_{m=1}^M$.

\subsection{Simplifying Game Structures} \label{sec:Simplifying-Game-Structures}

Equation~\eqref{eq:LQ-Advantage} requires a  parametric model of the functions $\bVhat^\theta$, $\bmu^{\theta}$,$\{\bP^\theta_{i}, \bPsi_i \}_{i\in\mfN}$, which results in potentially a very large parameter space and in principle result in requiring many training steps. In many cases, however, the structure of the  game can significantly reduce the dimension of the parameter space and leads to easily learnable model structures. The following subsections enumerate these typical simiplications.

\subsubsection*{Label Invariance}

Many games have symmetric players, and hence are invariant to a permutation of the label of players. Such \textit{label invariance} implies that each agent-$i$ does not differentiate among other game participants and the agent's reward functional is independent of any reordering of all other agents' states and/or actions.

% and may reorder the actions and portions of the game state tied to other agents without any change to agent-$i$'s reward functional.

% and the agent's reward functional is independent of any reordering of (i)  the actions of other agents, and (ii) all other agents' states.

% and the agent's reward functional is independent of any reordering of all other agents' states and/or actions.
 
%  [x_0, x_i, x_{-i} ]

% x_{-i} = {x_1,\dots, x_{i-1},     x_{i+1}, x_{i+2}, \dots, x_{N} }

More formally, we assume that for an arbitrary agent-$i$, the game's state can be represented as $x=(x_{0}, x_i, x_{-i} )$, where $x_0$ represents the part of the game state not belonging to any agent, $x_i$ represents the portion of the game state belonging to agent-$i$ and $x_{-i} = \{x_j \}_{j\in \mfN/\{i\}}$ represents the part of the game state belonging to other agents. Next, let  $\Lambda$ denote the set of permutations over sets of $N-1$ indices, where for each $\lambda\in\Lambda$, we express the permutation of a collection as $\lambda(\{ y_j \}_{j=1}^{N-1}) = \{ y_{k(j)} \}_{j=1}^{N-1}$, where $k:\{1,\dots,N-1\} \rightarrow \{1,\dots,N-1\}$ is a one-to-one and onto map from the indices of the collection into itself.

Label invariance is equivalent to the assumption that for any $\lambda\in\Lambda$, each agent's reward function satisfies
\begin{equation}
	r_i \left(
	x_0 , x_i , \lambda(x_{-i})
	\, ; \,
	u_i , \lambda(u_{-i})
	\right)
	=
	r_i \left(
	x_0 , x_i , x_{-i}
	\, ; \,
	u_i , u_{-i}
	\right)
	\;.
\end{equation}
With such label invariance,  the form of the linear quadratic expansion of the advantage function in~\eqref{eq:LQ-Advantage} simplifies. Assuming that $d_j = d$, for all $j\in\mfN$, independent label invariance in only the actions of agents requires $\Ahat_i$ to have the simplified form
\begin{equation} \label{eq:Invariant-Advantage}
	\begin{aligned}
		\Ahat_i^{\theta} (x;\bu) =
		&- \left\| u_i - \mu_i^{\theta}(x) \right\|_{P^{\theta}_{11,i}(x)}^{2}
		- \, \sum_{j\in\mfN/\{i\}}
		\left\langle  \; \left(u_i - \mu_i(x)^{\theta}\right) \, , \,
		 \left(u_j - \mu_j^{\theta}(x)\right) \; \right\rangle_{P^{\theta}_{12,i}(x)}
		\\
		&- \sum_{j\in\mfN/\{i\}} \lVert u_j - \mu_j^{\theta}(x) \rVert^{2}_{P^{\theta}_{22,i}(x)}
		\,+ \sum_{j\in\mfN/\{i\}} \left(u_j - \mu_j^{\theta}(x)\right)^\intercal \bpsi^{\theta}(x)\,,
	\end{aligned}
\end{equation}
for all $ i\in\mfN$,
where we use the notation $\|z\|^2_{M} = z^{\intercal} M z$ and $\langle y, z \rangle_M = y^{\intercal} M z$ for appropriately sized matrices $M$. 
% In equation~\eqref{eq:Invariant-Advantage},  we define for each $i$ and $\theta$ the functions $p^{\theta}_{1,2,i},p^{\theta}_{22,i}: \X \rightarrow \R^{d \times d}$ and $\bpsi^{\theta}: \X \rightarrow \R^{d}$. 
The functional form of \eqref{eq:Invariant-Advantage} allows us to drastically reduce the size of the matrices being modelled by an order of $N^2$.

To impose label invariance on states, we require  permutation invariance on the inputs the function approximations $\bVhat^\theta$, $\bmu^{\theta}$, $\{\bP_i,,\bpsi^{\theta}\}_{i\in\mfN}$. \cite{zaheer2017deep} provide necessary and sufficient conditions on neural network structures to be  permutation invariant. 
This necessary and sufficient structure is defined as follows. 
Let $\phi:\R^{n_1} \rightarrow \R^{n_2}$ and $\sigma:\R^{n_2} \rightarrow \R^{n_3}$ be two arbitrary functions. From these functions, let $f_{inv}:\R^{J \times n_1}\rightarrow\R^{n_3}$ be the composition of these functions, such that
\begin{equation} \label{eq:Invariant-Network-Structure}
	f_{\text{inv}}(z) = \sigma\left( \sum_{j=1}^J\phi( z_j ) \,\right).
\end{equation}
It is clear that $f_{inv}$ constructed in this manner is  invariant to the reordering of the components of $z$. Equation~\eqref{eq:Invariant-Network-Structure} may be interpreted as  a layer which aggregates the all dimensions of the inputs (which will corresponding to the state of all agents), through $\phi$, and a layer that transforms the aggregate result to the output, through $\sigma$. We assume further that $\phi$ and $\sigma$ are both neural networks with appropriate input and output dimension. This structure can also be embedded as an input later inside of a more complex neural network.

\subsubsection*{Identical Preferences}

It is quite common that  the admissible actions of all agents are identical,  i.e., $\U_i=\U$, $\forall i\in\mfN$, and agents have  homogeneous   objectives, or large sub-populations of agents have homogeneous objectives.  Thus far, we allowed agents to assign different performance metrics, and the variations are show through the set of rewards and discount rates, $\{ r_i, \gamma_i \}_{i\in\mfN}$. If agents have identical preferences, then we simply need to assume $r_i(x;\bu)=r(x;\bu)$ and $\gamma_i = \gamma$ for all $i\in\mfN$. By the definition of total discounted reward, state-action value function, and value function, identical preferences and admissible actions imply that $\Ri$, $Q_i$ and $V_i$ are independent of $i$.

In addition, the assumption of identical preferences, combined with the assumption of label invariance can further reduce the parametrization of the advantage function. Under this additional assumption we have that $\Vhat_i,\Ahat_i$ must be identical for all $i$, which reduces  modelling of all of the $\Vhat_i^\theta$, $\mu_i^{\theta}$,$\bP_i^\theta$,$\bpsi^{\theta}$ to modelling these for a single $i$. This further reduces the number of functions that must be modeled by an order of $N$. The combined effect of label invariance and identical preferences has a compounding effect which can have a large impact on the modelling task, particularily when considering large populations of players.

\begin{remark}[Sub-population Invariance and Preferences]
We can also consider cases where label and preference invariance   occur within sub-population of agents, rather than across the entire population. For example, in games in which some agents may cooperate with other agents, we can assume that agents are indifferent to re-labeling of cooperators and non-cooperators separately. Similarly, we can consider cases in which groups of agents share the same performance metrics. Such situations, among others, lead to modelling simplifications similar to equation~\eqref{eq:Invariant-Advantage} and simplifying neural network structures can be developed. In the interest of space, we do not develop further examples simplifying examples, nor do we claim the list we provide is exhaustive as one can easily imagine a multitude of other almost symmetric cases that can be of interest.
\end{remark}

\section{Implementation of Nash Actor-Critic Algorithm}
\label{sec:Implementation-Details-Section}
With the locally linear-quadratic form of the advantage function, and the simplifying assumptions outlined in the previous section, we can now minimize the objective~\eqref{eq:Empirical-Nash-Gap}, which reduces to the sum over \eqref{eqn:LossLocallyLQNash}, over the parameters $\theta$ through an iterative optimization and sampling scheme. One could in principle apply a simple stochastic gradient descent method using back-propagation on the appropriate loss function. Instead, we propose an actor-critic style algorithm to increase stability and efficiency of the algorithm. Actor-critic methods (see e.g.~\cite{konda2000actor}) have been shown to provide faster and more stable convergence of reinforcement learning methods towards their optima, and our model lends itself naturally to  such methods.

The decomposition in Equation~\eqref{eq:Q-Function-Decomposition} allows us to model the value function $\Vhat$ independently from other components. Therefore, we  employ an actor-critic  update rule  to minimize the loss function~\eqref{eqn:LossLocallyLQNash} by separating the parameter set  $\theta=(\theta_{V},\theta_{A})$, where $\theta_V$ represents the parameter set for modelling $\Vhat^{\theta_V}$ and $\theta_A$ represents the parameter set used for modeling $\bAhat^{\theta_A}$. Our proposed actor-critic algorithm updates these parameters by minimizing the total loss 
\begin{subequations}
\begin{equation}
\tfrac{1}{M}\sum_{m=1}^M \Lhat\left(y_m,\theta_V,\theta_A\right),
\end{equation}
where the individual sample loss corresponding to the error in the Nash-Bellman equation, after already solving for the Nash-equilibria, is 
\begin{equation}
	\Lhat(y_m,\theta_V,\theta_A) = \left\lVert
	\bVhat^{\theta_V}(x_m) +
	\bAhat^{\theta_A} ( x_{m} ; \bu_{m} )
	- \br(x_{m};\bu_{m}) - \gamma_i \, \bVhat^{\theta_V}(x_m^{\prime})
\right\rVert^2,
\end{equation}%
\label{eqn:NashA2C}
\end{subequations}%
with $\bu_{m}$, $y_m=\left(x_m,\bu_{m}, x_m^{\prime} \right)$, and  we minimize the loss by alternating between  minimization in the variables $\theta_A$ and $\theta_V$.

Algorithm \ref{alg:Deep-LQ-AC} provides an outline of the actor-critic procedure for our optimization problem. We include  a replay buffer and employ mini-batching. A replay buffer is a collection of previously experienced transition tuples of the form $y_t = (x_{t-1},\bu,x_t)$ representing the previous state of the system, the action taken in that state, the resulting state of the system, and the reward during the transition. We randomly sample a mini-batch from the replay buffer to update the model parameters using SGD. The algorithm also uses a na\"ive Gaussian exploration policy, although it may be  replaced by any other action space exploration method.
During the optimization steps over $\theta_V$ and $\theta_A$, we use stochastic gradient descent, or any other adaptive optimization methods.
\begin{algorithm}[t]
	\Input{\# Episodes $B>0$, Minibatch Size $\hat{M}>0$, \# of Game Steps $N$, Exploration Noise $\{\sigma_b\}_{b=1}^B > 0$ } \;
	\Initialize{Replay Buffer $\D$, Parameters $(\theta_A,\theta_V)$ } \;
	\For{Episode $b \leftarrow 1$ \KwTo $B$}{
		Reset Simulation, get initial state $x_0$. \; \\
	    \For{Game steps $t\leftarrow 1$ \KwTo $N$ }
	    {	
	        \KwSty{Select} actions $\bu \leftarrow \bmu^{\theta_A}(x) + \bm{\epsilon} \; , \bm{\epsilon} \sim \mvn(0,\sigma_b I)$.\; \\
	        \KwSty{Observe} state transition $y_t = (x_{t-1},\bu,x_t)$ from game. \; \\
	        \KwSty{Store} $\D \leftarrow y_t = (x_{t-1},\bu,x_t)$ \; \\
	        \KwSty{Sample} $\Y=\{ y_i \}_{i=1}^{\hat{M}}$ randomly from $\D$ \; \\
	        \KwSty{Optimization Step} of $\frac{1}{\hat{M}+1} \sum_{y\in\Y\bigcup \{y_t\}} \Lhat(y,\theta_V,\theta_A)$ over $\theta_V$ \; \\
	        \KwSty{Optimization Step} of $\frac{1}{\hat{M}+1} \sum_{y\in\Y\bigcup \{y_t\}} \Lhat(y,\theta_V,\theta_A)$ over $\theta_A$ \; \\
	        }
	    } \;
	    \Return{ $(\theta_A,\theta_V)$}
	\caption{Nash-DQN Actor-Critic Algorithm}
	\label{alg:Deep-LQ-AC}
\end{algorithm}

\section{Experiments}
\label{sec:Experiments-Section}

We test our algorithm on a multi-agent game for statistical arbitrage in electronic exchanges. The game consists of agents trading a single asset with a stochastic  price process that is affected by their actions. A simpler version of this model has been studied in \cite{casgrain2018algorithmic,casgrain2020mean, neuman2021trading}. In these works, under various assumptions, the authors take a mean-field approach and show that the mean field optimal strategy provides an approximate Nash equilibrium for the finite-player case.

In our setting, an arbitrary agent-$i$, $i\in\mfN$, may buy or sell assets at a rate of $\nu_{i,t} \in \R$ during the trading horizon $t\in\mfT:=[0, T]$. At $t=T$,  agents must liquidate their holdings, any remaining inventory will be subjected to a terminal penalty. Agents may change their rate of trading at discrete specific points in time $t_1, t_2, \dots, t_M$ (decision points). The rate of trading is assumed to be constant between any two decision points and thus the total amount of assets traded during the period $(t_m, t_{m+1})$ is given by $\nu_{i,t_m}^T = (t_{m+1} - t_m) \nu_{i,t_0}$. Each agent-$i$ keeps track of their inventory $q_{i,t_m} = q_{i,0} + \int_0^{t_m} \nu_{i,t} \, dt$. Inventories are not visible to other agents however the total order flow, i.e. $\sum_{i\in\mfN}(q_{i,t} - q_{i,0})$, is. For ease of notation, we assume $t \in \{ t_1, t_2, \dots, t_M \}$. 
We assume the asset price process $S_t$  evolves according to the following system of continuous time dynamics:
\begin{align}
	dS_t &= (\mu(S_t) + g(\bnu_t)) \, dt + dY_t + \sigma \,dW_t, \\
	dY_t &= -\rho\,Y_t\,dt + \gamma \, h(\bnu_t) \, dt,
	\label{eq:smpl-dynamics}
\end{align}
with initial condition $S_0$, variance $\sigma > 0$, and Brownian motion $(W_t)_{t\ge0}$. Furthermore, we denote the vector valued process of trading rates by $\bnu = (\left( \nu_{i,t} \right)_{i\in\mfN})_{t\ge0}$ and denote by $\nubar_t = \sum_{i\in\mfN} \nu_{i,t}$ the total trading rate at time $t$. The above form assumes the cumulative trades of all agents induce both a transient impact through $Y:=(Y_t)_{t\ge0}$ and a permanent impact through $g$.

The functions $\mu$ and $g$ represents the mean process and permanent price impact respectively. For the experiments conducted here, we assume the process mean-reverts so that $\mu(S) = \kappa(\theta - S)$, where $\kappa>0$ represents the mean-reversion rate and $\theta$ the mean-reversion level. Moreover, we assume a linear permanent price impact so that $g(\bnu) = \eta \, \nubar $. The cumulative transient price impact is assumed to be square root, i.e. $h(\bnu) = sign(\nubar)\sqrt{\nubar}$. Under these model assumptions, the continuous dynamics may be written
\begin{align}
	dS_t &= \left(\kappa(\theta - S_t) + \eta \,\nubar_t\right)\, dt + dY_t  + \sigma W_t \\
	dY_t &= -\rho\,Y_t\,dt + \gamma \, sign(\nubar_t) \, \sqrt{\nubar_t} \, dt
	\label{eq:full-dynamics}
\end{align}

% An analytical solution to this particular setup is studied \cite{neuman2021trading}, however the authors assume a linear transient price impact which allows the problem to be tractable.

In addition, each agent pays a transaction cost proportional to the amount they buy or sell during each time period. This represents the cost incurred by the agents from ``walking the limit-order-book'' and is often dependent on the liquidity of the asset. Agents keep track of their total cash from trading and we denote the corresponding process by $X_i:=(X_{i,t})_{t\ge0}$ where $X_{i,t} = - \int_0^T \nu_{i,t} \left( S_{t} + b_1 \nu_{i,t} \right) \, dt$, and $b_1 > 0$ is the transaction cost constant.

The agent's objective is to maximize the sum of (i) total cash they possess by time $T$, (ii) excess exposure at time $T$, and (iii) a penalty for risk taking. We express agent-$i$'s objective (total expected reward) as
\begin{equation} \label{eq:Trading-Objective}
	\Ri:=
	\E\left[ X_{i,T} + q_{i,T} \left( S_{T} - b_2 \, q_{i,T} \right)
	- b_3 \int_0^T q_{i,t}^2 \, dt \right],
\end{equation}
where $b_2,b_3>0$. In Equation~\eqref{eq:Trading-Objective}, the second term serves as a cost of instantaneously liquidating the inventory at time $T$ and the last term serves as a penalty for taking on excess risk proportional to the square of the holdings at each time period -- the so called urgency penalty. In this objective function, the effect of all agent's trading actions appears implicitly through the dynamics of $S$, and through its effect on the cash process $X_{i}$. This particular form of objective assumes that agents have identical preferences which are invariant to agent relabeling\footnote{We could extend this to include the case of sub-populations with homogeneous preferences, but that are heterogeneous across sub-populations.}. Hence, we may employ the techniques discussed in \Cref{sec:Simplifying-Game-Structures} to simplify the form of the advantage function $\Ahat$. In our example, we model each component of the advantage function $\Ahat$ with neural networks that includes a permutation invariant layer. 

Our experiments assume a total of five agents over a time horizon of five hours ($T=5$) and ten decision points ($M=10$) of equal duration.

\subsection{Features}
\label{sec:features}
We use the following features to represent the state $x_t$ of the environment at time $t$:\\
\textbf{Price ($S_t$): } Scalar representing the current price of the asset, \\
\textbf{Time ($t$): } Scalar representing the current time step the agent is at in the time horizon, \\
\textbf{Total order flow ($\sum_{i \in \mfN} q_{i,0} - q_{i,t} $): } Scalar representing the total order from all agents since the beginning of the trading horizon, and \\
\textbf{Cumulative transient price impact ($Y_t$): } Scalar representing the current level of transient price impact effecting the price.

All features are assumed to be non-label invariant. 
%\footnote{We could instead allow for individual agent's order flow, which would then be considered label invariant.}

\subsection{Network Details}

The network structure for the advantage function approximation $\bAhat^{\theta_A}$ consists of two network components: (i) a permutation invariant layer that feeds into (ii) a main network layer. The input of the permutation invariant layer are the label invariant features. This layer, as described in \Cref{sec:Simplifying-Game-Structures}, is a fully connected neural network with three hidden layers each containing 20 nodes. Layers are connected by SiLU activation functions \cite{elfwing2018sigmoid}.  We then combine the output of this permutation invariant later with the non-label invariant features and together they form the inputs to the main network. The main network comprises of four hidden layers with 32 nodes each.  The outputs of this main network are the parameters $\bmu^{\theta}$ and $\{\bP^\theta_{i}, \bPsi_i \}_{i\in\mfN}$, of the approximated advantage function defined in \Cref{sec:Locally-Linear-Quadratic-Algortithm}. These parameters fully specify  the value of the advantage function. 

The network structure for the value function approximation  $\hat{V}^{\theta_V}$ contains four hidden layers with 32 nodes each. This network takes the features from all states described in \Cref{sec:features} and outputs the approximate value function for all agents.

We use the Adam optimizer \cite{loshchilov2018decoupled} with mini-batches and a weight decay of 0.001 to optimize the loss functions defined in \Cref{sec:Implementation-Details-Section}. Mini-batch sizes are set to ten full episodes executed on the period [0, T]. Learning rates are set to 0.003 and are held constant throughout training, modified only by the optimizer's weight decay. Training is performed over a maximum of 20,000 iterations, with a early stopping criteria of no improvement in the loss in the last 3,000 iterations.

\subsection{Baseline - Fictitious Play}

Fictitious Play (FP), first introduced in \cite{brown1951iterative}, is a classical method of determining the Nash Equilibrium of multi-agent games. It has been shown to converge in the two player case for zero-sum games \cite{brown1951iterative}, potential games \cite{monderer1996potential}, $2\times N$ games with generic payoffs \cite{berger2005fictitious}, and is often used as a basis for many modern methods \cite{heinrich2015fictitious}. In general, FP assumes each player follows a stationary mixed strategy that is updated at each iteration by taking the best response to the empirical average of the opponents' previous strategies. In our experiments we assume identical agents and thus computing the optimal response for a single agent at each iteration is sufficient. Specifically, let ${\bu^F_i(x)}_{i\in \mfN}$ be the optimal FP strategies for each agent which we use the algorithm defined in \Cref{alg:Fic-Play} to obtain.

\begin{algorithm}[t]
\setcounter{AlgoLine}{0}
	\Initialize{$\{\bu^F_i(x)\}_{i\in \mfN, i \neq 0}$} \;
	\For{Iteration $b \leftarrow 1$ \KwTo $B$}{
	    \KwSty{Optimize} $u^F_0(x)$ assuming others act according to $\{\bu^F_i(x)\}_{i\in \mfN, i \neq 0}$. 
	    \label{alg:optimize}
	    \\
	    \KwSty{Assign} all other agents to this iterations optimal: $\{\bu^F_i(x)\}_{i\in \mfN, i \neq 0} \leftarrow u^F_0(x)$
	}
	\Return{ $u^F_0(x)$}
	\caption{Fictitious Play Algorithm}
	\label{alg:Fic-Play}
\end{algorithm}

To obtain the optimal response of agent-$i$ against other agents' strategies in line \ref{alg:optimize} of Algorithm \ref{alg:Fic-Play} we use Deep Deterministic Policy Gradient (DDPG) \cite{lillicrap2015continuous}. DDPG uses a neural network to represent the optimal policy and the state-action value function. As with our Nash-DQN approach, this approach has no knowledge of the agent's reward function nor the transition dynamics, and hence  provides a fair comparison for our Nash-DQN. Due, however, to the nature of DDPG combined with FP it  is highly data and training inefficient as each FP iteration requires the optimization of the DDPG policy -- and each iteration of DDPG uses approximately the same amount of resources as a full training cycle of our Nash-DQN method.  

%As comparisons, we will use two different methods of policy optimization: (i) Deep Deterministic Policy Gradient \cite{lillicrap2015continuous}, and (ii) reward maximization using backprop. The first method uses a neural network to represent the optimal policy and the state-action value function and is then optimized iteratively. As this method uses a neural network approximation of the state-action value function using only rewards obtained from the simulator, no knowledge of any agent's reward functions nor transition dynamics are leaked to the agent and thus provides a fair comparison for the Nash-DQN. The second method however assumes explicit knowledge of both the agent's reward function and the system's transition dynamics by optimizing the policy to maximize the rewards by stochastic gradient descent and thus serves as a theoretical optimum for the Nash-DQN to achieve.

\subsection{Optimization Improvements}

\begin{wrapfigure}{r}{0.4\textwidth}
\centering
\includegraphics[width=\linewidth]{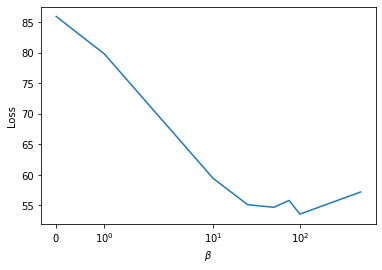}
\caption{Average loss for different $\beta$ values. Optimal value is $\beta = 100$.}
\label{fig:penalty_loss}
\end{wrapfigure}
We find that directly minimizing \Cref{eqn:NashA2C} sometimes produce inaccuracies in the final policy due to a difference in magnitudes of the learned parameters used in the advantage function \Cref{eq:Invariant-Advantage}, specifically the linear term $\bpsi^{\theta}(x)$. Adding in the $L^1$ regularization term 
\begin{equation}
	\tilde{L}(y_m,\theta_V,\theta_A) := \Lhat(y_m,\theta_V,\theta_A) + \beta \, \| \bpsi^{\theta}(x) \|
\end{equation}%
improves the performance significantly.

The optimal value for the hyperparameter $\beta$ is determined by minimizing the average loss \Cref{eqn:NashA2C} computed over 1,000 randomly generated simulations. \Cref{fig:penalty_loss} indicates an value of $\beta = 100$ is optimal.

\subsection{Results}

We use the set of model parameters shown in \Cref{tab:ModelParameters} in our analysis.
\begin{table}[h!]
  \centering
  \caption{Asset price process, price impact, and risk preference parameters.}
    \begin{tabular}{rrrrrrrrrr}
    \toprule
    \toprule
    \multicolumn{1}{l}{$\kappa$} & \multicolumn{1}{l}{$\theta$} & \multicolumn{1}{l}{$\sigma$} & 
    \multicolumn{1}{l}{$\gamma$} & \multicolumn{1}{l}{$\rho$} &
    \multicolumn{1}{l}{$\eta$} &
    \multicolumn{1}{l}{$b_1$} & $b_2$ & \multicolumn{1}{l}{$b_3$}& $\Delta T$ \\
    \midrule
    0.1   & 10    & 0.01 & 0.02 & 0.5 & 0.05   & 0.1 & 0.1 & 0 & 0.5\\
    \bottomrule
    \bottomrule
    \end{tabular}%
  \label{tab:ModelParameters}%
\end{table}%

\begin{figure}[t!]
\centering
\begin{subfigure}{.32\textwidth}
  \centering
  \includegraphics[width=1\linewidth]{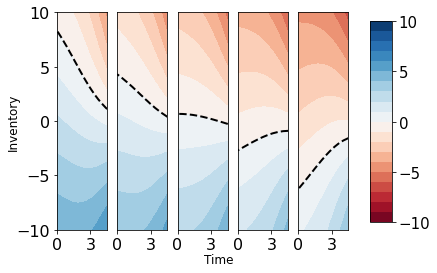}
  \caption{NashDQN, $Y_{t}=-0.2$}
\end{subfigure}%
\begin{subfigure}{.32\textwidth}
  \centering
  \includegraphics[width=1\linewidth]{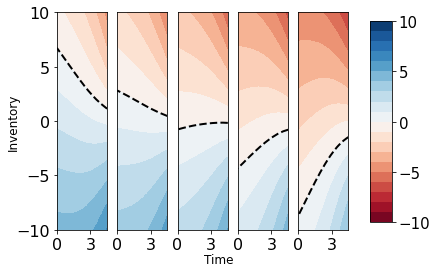}
  \caption{NashDQN, $Y_{t}=0$}
\end{subfigure}
\begin{subfigure}{.32\textwidth}
  \centering
  \includegraphics[width=1\linewidth]{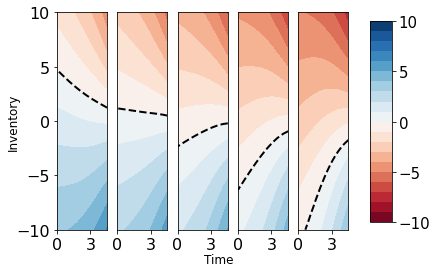}
  \caption{NashDQN, $Y_{t}=0.2$}
\end{subfigure}

\begin{subfigure}{.32\textwidth}
  \centering
  \includegraphics[width=1\linewidth]{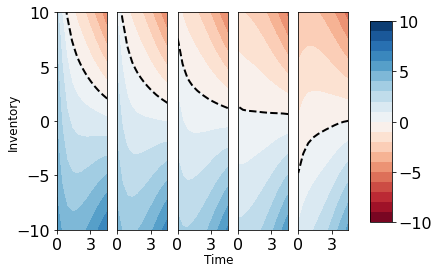}
  \caption{Fic Play, $Y_{t}=-0.2$}
\end{subfigure}
\begin{subfigure}{.32\textwidth}
  \centering
  \includegraphics[width=1\linewidth]{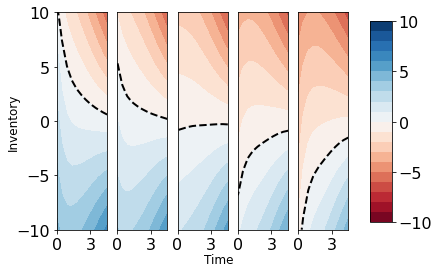}
  \caption{Fic Play, $Y_{t}=0$}
\end{subfigure}
\begin{subfigure}{.32\textwidth}
  \centering
  \includegraphics[width=1\linewidth]{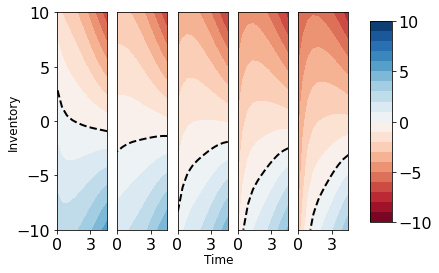}
  \caption{Fic Play, $Y_{t}=0.2$}
\end{subfigure}
\caption{Optimal trade execution heatmaps of the Nash-DQN and Fictitious Play policies as a function of time, inventory, price, and cumulative transient impact ($Y_t$), other features were assumed to be equal to zero. Within each panel,  subplots corresponds to  price levels $\$9.50,\$9.75,\dots,\$10.50$ from left to right. The dotted line shows the threshold where the agent switches from buying to selling.  
\label{fig:PolicyHeatmaps}}
\end{figure}

\begin{figure}[t!]
\centering
\begin{subfigure}{.45\textwidth}
  \centering
  \includegraphics[width=1\linewidth]{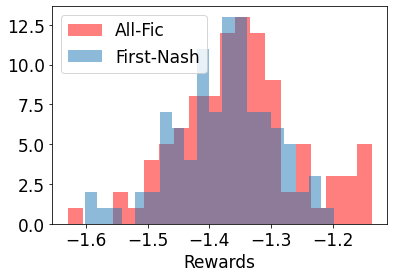}
  \caption{All Fic. Deviation}
\end{subfigure}%
\begin{subfigure}{.45\textwidth}
  \centering
  \includegraphics[width=1\linewidth]{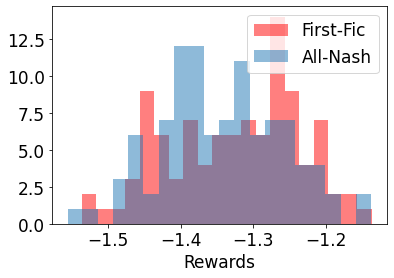}
  \caption{All NashDQN Deviation}
\end{subfigure}
\caption{Results of deviating from the Fictitious Play policy (i) and the Nash-DQN Policy (ii) when taking the total rewards obtained through randomly initialized simulated environments. Each data point represents the average of 1,000 simulations, repeated 100 times.
\label{fig:HistoResults}}
\end{figure}

\Cref{fig:PolicyHeatmaps} shows the optimal policies obtained through our Nash-DQN method and the FP method. We can see that despite being significantly more data efficient, the optimal policies are essentially the same. To evaluate the performance of the Nash-DQN method, we considering the following two scenarios: (i) Agent 1 following Nash-DQN and all other agents FP vs. all agents following FP, (ii) Agent 1 following FP and all other agents Nash-DQN vs all agents following Nash-DQN. Next, we apply these policies to 1,000 paths using the simulated environment. Finally, we average total rewards, repeat the whole exercise 100 times, and plot the resulting distributions shown in \Cref{fig:HistoResults}. From the figures, it appears there is no noticeable discernable difference in the distributions. Indeed, the null hypothesis that the means of the results from FP and Nash-DQN differ cannot be rejected at the 5\% level.

\section{Conclusions}

Here we present a computationally tractable  reinforcement leanring framework for multi-agent (stochastic) games. Our approach utilizes function approximations after decomposing the collection of agents' state-action value functions into the individual value functions and their advantage functions. Further, we approximate the advantage function in  a locally linear-quadratic form and use neural-net architectures to approximate both the value  and  advantage function. Typical symmetries in  games allow us to use permutation invariant neural-nets, motivated by the Arnold-Kolmogorov representation theorem, to reduce the dimensionality of the parameter space. Finally, we  develop an actor-critic paradigm to estimate parameters and apply our approach to an important application in electronic trading. Our approach is more data efficient than conventional FP policies, and is applicable to large number of players and continuous state-action spaces.

There are a number of doors left open for exploration including extending our approach to account for latent factors driving the environment, and when the state of all agents are partially (or completely) hidden from any individual agent. As well, our approach can be easily applied to mean-field games which correspond to the infinite population limit of stochastic games that have interactions where any individual agent has only an infinitesimal contribution to the state dynamics.

\bibliographystyle{siamplain}
\bibliography{nash_biblio}
% \nocite{*}

\end{document}